\documentclass[10pt,twocolumn,letterpaper]{article}

\usepackage{wacv}              % To produce the CAMERA-READY version
\usepackage{graphicx}
\usepackage{amsmath}
\usepackage{amssymb}
\usepackage{booktabs}
\usepackage{pgfplots}
\usepackage{tikz}
\usepackage{multirow}
\usepackage{subcaption}
\usepackage{pifont}% http://ctan.org/pkg/pifont
\usepackage{algorithm2e}
\newcommand{\cmark}{\ding{51}}%
\newcommand{\xmark}{\ding{55}}%

\newcommand{\x}{\boldsymbol{x}}
\newcommand{\I}{\boldsymbol{I}}
\newcommand{\N}{\mathcal{N}}
\newcommand{\bs}[1]{\boldsymbol{#1}}
\newcommand{\nbr}[1]{\left(#1\right)}
\newcommand{\sbr}[1]{\left[#1\right]}

\usepackage[pagebackref,breaklinks,colorlinks]{hyperref}

\usepackage[capitalize]{cleveref}
\crefname{section}{Sec.}{Secs.}
\Crefname{section}{Section}{Sections}
\Crefname{table}{Table}{Tables}
\crefname{table}{Tab.}{Tabs.}

\def\wacvPaperID{} % *** Enter the WACV Paper ID here
\def\confName{WACV}
\def\confYear{2025}

\begin{document}

\title{AnoFPDM: Anomaly Detection with Forward Process of Diffusion Models for Brain MRI}

\author{Yiming Che$^{1,2}$, Fazle Rafsani$^{1,2}$, Jay Shah$^{1,2}$, Md Mahfuzur Rahman Siddiquee$^{1,2}$, Teresa Wu$^{1,2}$\\
$^1$Arizona State University\\
$^2$ASU-Mayo Center for Innovative Imaging\\
% {\tt\small email@institution1.edu}
}
\maketitle

%%%%%%%%% ABSTRACT
\begin{abstract}
  Weakly-supervised diffusion models in anomaly segmentation, which leverage image-level labels and bypass the need for pixel-level labels during training, have shown superior performance over unsupervised methods, offering a cost-effective alternative. Traditional methods that rely on iterative image reconstruction are not fully weakly-supervised due to their dependence on costly pixel-level labels for hyperparameter tuning in inference. To address this issue, we introduce Anomaly Detection with Forward Process of Diffusion Models (AnoFPDM), a fully weakly-supervised framework that operates without image reconstruction and eliminates the need for pixel-level labels in hyperparameter tuning. By using the unguided forward process as a reference, AnoFPDM dynamically selects hyperparameters such as noise scale and segmentation threshold for each input. We improve anomaly segmentation by aggregating anomaly maps from each step of the guided forward process, which strengthens the signal of anomalous regions in the aggregated anomaly map. Our framework demonstrates competitive performance on the BraTS21 and ATLAS v2.0 datasets. Code is available at \url{https://github.com/SoloChe/AnoFPDM}.
\end{abstract}

%%%%%%%%% BODY TEXT
\section{Introduction}
\label{sec:intro}
Anomaly detection is a critical task in the medical domain, aiding radiologists in diagnostics and subsequent decision-making. Traditionally, this process requires extensive pixel-level annotations, which are not only costly but also subject to human annotator bias \cite{tajbakhsh2020embracing}. Although supervised non-generative models have achieved state-of-the-art performance \cite{isensee2021nnu}, they still require a substantial amount of pixel-level labels, which are both expensive and labor-intensive to acquire. In contrast, weakly-supervised generative models have attracted attention for their ability to utilize image-level labels, categorizing images simply as healthy or unhealthy, which are more cost-effective and less prone to bias. These models have shown superior performance in anomaly segmentation tasks compared to unsupervised models \cite{wolleb2022diffusion,sanchez2022healthy,hu2023conditional}.

Among generative models, diffusion models (DMs) \cite{ho2020denoising,song2020denoising,song2020score} are often favored due to their superior image generation capabilities compared to other models such as generative adversarial networks (GANs) \cite{goodfellow2014generative} and variational autoencoders (VAEs) \cite{kingma2013auto}.  However, despite the weakly-supervised nature of current approaches, they still significantly rely on pixel-level labels, particularly during the hyperparameter tuning phase. For methods based on DMs, this tuning involves hyperparameters such as the \textbf{guidance strength}, fixed amount of noise (\textbf{noise scale}) added to the input images, and \textbf{fixed threshold} for generating the predicted pixel-level labels from anomaly map. This reliance on pixel-level labels for tuning reintroduces the associated costs and biases, undermining the potential benefits of weakly-supervised learning.

In weakly-supervised anomaly detection with DMs, the model is trained on both healthy and unhealthy samples, as proposed in \cite{dhariwal2021diffusion,ho2021classifier}. This training approach enables the model to learn the distribution of healthy and unhealthy samples, and subsequently map unhealthy inputs to the healthy distribution through a partial diffusion process. During this process, the partially noised unhealthy inputs are reconstructed as healthy by the iterative sampling process (backward process) of DMs, which aims to minimize the presence of anomalies in reconstructed unhealthy inputs while preserving other healthy regions as intact as possible. However, the reconstruction error tends to be accumulated due to the iterative sampling. The crucial step of anomaly detection involves thresholding the difference between the original inputs and their reconstructed counterparts, i.e., anomaly map, to obtain the predicted pixel-level labels. Prior to this, the hyperparameters are selected using grid search with pixel-level labels to optimize metrics such as the DICE score. The noise scale manages the level of noise added to the inputs, balancing the removal of anomalies against the preservation of healthy regions in the reconstructed inputs. If excessive noise is applied, the healthy regions may be overly distorted, rendering accurate segmentation impractical. Conversely, insufficient noise might fail to eliminate the anomalies. Given that alterations to the healthy regions are inevitable, it is crucial to carefully select the noise scale. This selection should aim for a `sweet spot' that optimally balances the removal of anomalous regions with the preservation of healthy areas. The guidance strength primarily mediates between the Frechet Inception Distance (FID) and the Inception Score (IS), balancing the quality and diversity of the generated samples. Additionally, the fixed threshold is employed to identify pixels within the anomalous regions on the anomaly map. Given the variable signal strength of anomalous regions in the anomaly maps across different inputs, this fixed threshold might not always yield optimal results, suggesting the need for adaptable thresholding strategies to enhance model performance.

We introduce a fully weakly-supervised framework, AnoFPDM, which leverages diffusion models equipped with classifier-free guidance \cite{ho2021classifier} to obviate the need for pixel-level labels during hyperparameter tuning. During the training phase, AnoFPDM adheres to the standard training protocol of classifier-free guidance as described in \cite{ho2021classifier}. In the inference stage, rather than employing the traditional iterative sampling process, i.e., image reconstruction, our framework capitalizes on the forward process of DMs. This involves progressively adding noise to the input images. Additionally, we incorporate a crucial step by obtaining both healthy-guided predictions (HGPs) and unguided predictions (UGPs) of original inputs through a one-step mapping, referred to as the healthy-guided forward process (HFP) and unguided forward process (UFP), respectively. We employ the aggregation of sub-anomaly maps (SAMs), the difference between the HGPs and original inputs at each step, to enhance the signal strength of anomalous regions, thereby improving the segmentation performance. Each SAM captures signals from anomalous regions at a specific frequency. The signal strength of anomalous regions is consolidated by aggregating anomalous signals from different frequencies, while reducing the mistakenly detected healthy signals in the aggregated anomaly map. This reduction is effective because signals from healthy regions are less consistent across frequencies. Traditional methods, although accurate in detecting anomalies, often fail to suppress healthy signals, resulting in relatively weaker anomaly signals.

There are three hyperparameters to be selected in AnoFPDM: the end step of aggregation, i.e., noise scale, threshold for the aggregated SAMs, and guidance strength. We propose a novel approach that dynamically selects the noise scale and threshold for each individual input, while maintaining a fixed guidance strength selection by utilizing UFP as a reference. Additionally, we obtain stronger signal strength of anomalous regions in the aggregated anomaly map (AAM). Overall, our contributions are as follows:
\begin{itemize}
  \item We propose a fully weakly-supervised anomaly detection framework without noised image reconstruction, AnoFPDM, that operates without the need of pixel-level labels for hyperparameter tuning.
  \item We introduce a novel dynamical threshold and noise scale selection and a novel guidance strength selection for DMs on weakly-supervised anomaly detection.
  \item We propose a novel aggregation strategy combined with dynamical noise scale selection to enhance the signal strength of anomalous regions.
\end{itemize}

\begin{figure}[!h]
  \centering
  \includegraphics[scale=0.40]{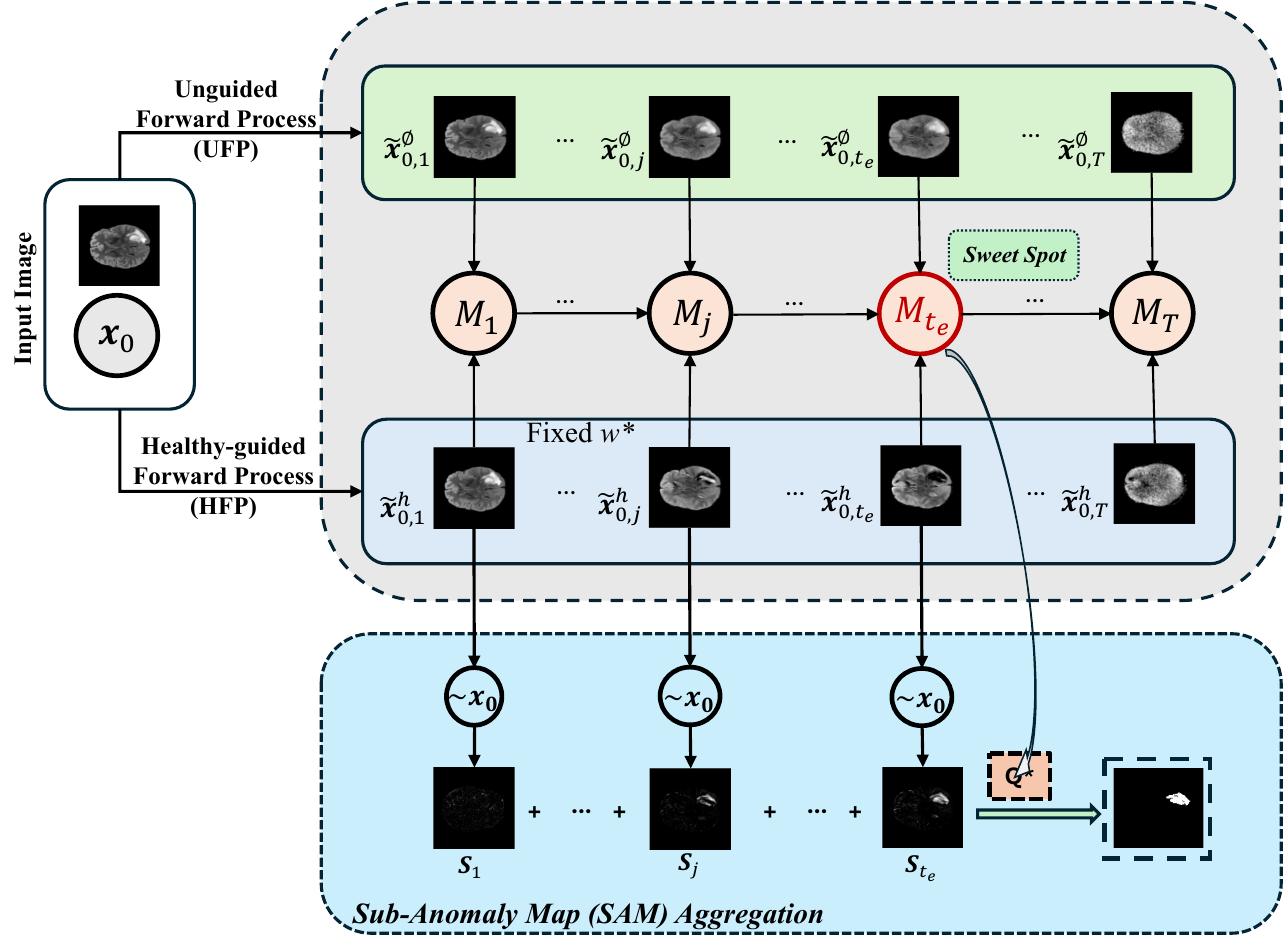}
  \caption{The diagram of the proposed method. Given the selected fixed guidance strength $w$, the noise scale $t_e$ and threshold $Q^*$ are dynamically selected for each input $\x_0$ according to the divergence $M_{t_e}$ between HFP and UFP calculated in each forward step. $\tilde{\x}_{0,t}^{h}$ and $\tilde{\x}_{0,t}^{\emptyset}$ are HGP and UGP respectively. The SAMs are aggregated to enhance the signal strength of anomalous regions.}
  \label{fig:diagram}
\end{figure}

\section{Related Work}
Prior to the advent of diffusion models (DMs), the field of anomaly detection was largely dominated by generative adversarial networks (GANs) \cite{goodfellow2014generative} and variational autoencoders (VAEs) \cite{kingma2013auto}. GANs, despite their powerful capabilities, often faced criticism due to unstable training dynamics, while VAEs were limited by their expressiveness in the latent space and a tendency to produce blurry reconstructions \cite{schlegl2019f, chen2020unsupervised, zimmerer2019unsupervised}. DMs emerged as a promising alternative, gaining recognition for their success in image synthesis \cite{song2020denoising,song2020score,ho2020denoising}. In unsupervised anomaly detection, innovations such as the unguided denoising diffusion probabilistic model (DDPM) with simplex noise have been notable, achieving significant outcomes \cite{wyatt2022anoddpm}. Enhancements in model performance have also been seen with the use of patched images \cite{behrendt2023patched} and masked images in both physical and frequency domains \cite{iqbal2023unsupervised}. Furthermore, the concept of latent diffusion models, which facilitate fast segmentation, has been effectively utilized \cite{rombach2022high, pinaya2022fast}. In the realm of weakly-supervised approaches, pioneering efforts such as those by Wolleb \etal \cite{wolleb2022diffusion} have utilized DDIM with classifier guidance \cite{dhariwal2021diffusion}. This approach has been paralleled by Sanchez et al. \cite{sanchez2022healthy}, who adopted DDIM with classifier-free guidance \cite{ho2021classifier}, similar to the dual diffusion implicit bridging (DDIB) approach \cite{su2022dual}. Hu \etal adopted DDIM with classifier guidance and employed the finite difference method to approximate the gradient of denoised inputs in terms of image-level labels during the iterative sampling process for segmentation \cite{hu2023conditional}.

\section{Background}
\subsection{Diffusion Models}
To approximate the unknown data distribution $q(\x)$, DMs perturb the data with increasing noise level and then learn to reverse it using a model $\bs{\epsilon_\theta}$, typically utilizing U-net like architectures \cite{ronneberger2015u}, parameterized by $\bs{\theta}$. The forward process of DDPM, perturbing the data, is fixed to a Markov chain with increasing variance schedule $\beta_1,...,\beta_T$ where $T$ is the total number of steps. The transition kernel from $\x_{t-1}$ to $\x_{t}$ is modeled as Gaussian 
\begin{equation}
    q(\x_t | \x_{t-1}) = \N\nbr{\x_t | \sqrt{1-\beta_t}\x_{t-1}, \beta_t\I}.
    \label{eq:transition}
\end{equation}
Hence, the transition from the input $\x_0$ to any arbitrary perturbed input $\x_t$ is still Gaussian $q(\x_t|\x_0) = \N\nbr{\sqrt{\bar{\alpha}_t}\x_{0}, (1-\bar{\alpha}_{t})\I}$. Then, we can sample from it by
\begin{equation}
    \x_t = \sqrt{\bar{\alpha}_t}\x_{0} + \sqrt{1-\bar{\alpha}_{t}} \bs{\epsilon}_t,
    \label{eq:ddpmforward}
\end{equation}
where $\bs{\epsilon}_t \sim \N\nbr{\bs{0}, \I}$, $\bar{\alpha}_t = \prod_{i=1}^t \alpha_i$ and $\alpha_t = 1-\beta_t$. 
The sampling process of DDPM, reversing the data from noise, is factorized as $p_\theta\nbr{\x_{0:T}} = p(\x_T)\prod_{t=1}^{T}p_\theta(\x_{t-1}|\x_t)$. It is an iterative transition from $\x_T$ to $\x_0$ with the learned distribution $p_\theta\nbr{\x_{t-1} | \x_t}$ which is the approximation of inference distribution $q(\x_{t-1} | \x_t, \x_0)$. The details can be found in \cite{ho2020denoising}. Then, we can iteratively sample from the learned distribution $p_\theta$ to obtain less noisy samples in closed form
\begin{equation}
    \x_{t-1} = \frac{1}{\sqrt{\bar{\alpha}_{t}}}\nbr{\x_t - \frac{\beta_t}{\sqrt{1-\bar{\alpha}_{t}}}\bs{\epsilon_\theta}(\x_t,t)} + \bs{\Sigma}_t\bs{z},
    \label{eq:ddpmsampling}
\end{equation}
where $\bs{z} \sim \N\nbr{\bs{0},\I}$ and $\bs{\Sigma}_t = \frac{1-\bar{\alpha}_{t-1}}{1-\bar{\alpha}_t}\beta_t\I$. Note that we can directly obtain the predicted input $\tilde{\x}_0$ from \cref{eq:ddpmforward} without the iterative sampling process as $\tilde{\x}_0 = \frac{\x_t - \sqrt{1-\bar{\alpha}_t} \bs{\epsilon_\theta}(\x_t, t)}{\sqrt{\bar{\alpha}_t}}$. However, the predicted input $\tilde{\x}_0$ will become less accurate as the noise level increases.

The variation DDIM \cite{song2020denoising} is non-Markovian, incorporating the input $\x_0$ into the forward process. The sampling process of DDIM is factorized as $q(\x_{1:T} | \x_0) = q(\x_{T} | \x_0)\prod_{t=2}^{T}q(\x_{t-1}|\x_t,\x_0)$. The forward transition from $\x_{t-1}$ to $\x_t$ can be obtained by Bayes's rule with the closed form $q(\x_{t+1}|\x_{t},\x_0) = \N\nbr{\sqrt{\bar{\alpha}_{t+1}}\x_0 + \sqrt{1-\bar{\alpha}_{t+1}-\sigma_t^2}\frac{\x_{t}-\sqrt{\bar{\alpha}_{t}}\x_0}{\sqrt{1-\bar{\alpha}_{t}}},\sigma_t^2\bs{I}}$ for $t\geq2$, where $\sigma_t$ is set to 0 to enable the deterministic perturbation. Then, we can perturb inputs iteratively by
\begin{equation}
  \x_{t+1} = \sqrt{\bar{\alpha}_{t+1}}\tilde{\x}_0 + \sqrt{1-\bar{\alpha}_{t+1}}\frac{\x_{t}-\sqrt{\bar{\alpha}_{t}}\tilde{\x}_0}{\sqrt{1-\bar{\alpha}_{t}}}.
  \label{eq:ddimforward}
\end{equation}
where $\x_1$ can be obtained by \cref{eq:ddpmforward}.

They both share the same objective function in training \cite{song2020denoising}. The training process is the minimization of KL divergence 
\begin{equation}
    KL\sbr{q | p_\theta} = \mathbb{E}_{\x_{1:T}\sim q(\x_{1:T}|\x_0)}\sbr{\log \frac{q(\x_{1:T}|\x_0) p_\theta(\x_0)}{p_\theta(\x_{0:T})}}.
\end{equation}
The objective is further simplified as $L_t = \left\lVert \bs{\epsilon}_t - \bs{\epsilon_\theta}\nbr{\sqrt{\bar{\alpha}_t}\x_{0} + \sqrt{1-\bar{\alpha}_{t}} \bs{\epsilon},t}\right\rVert$. The model functions primarily as a noise predictor, learning to add noise in a way that inversely teaches it about the underlying data distribution $q(\x)$.

\subsection{Classifier-free Guidance}
We employ the classifier-free guidance \cite{ho2021classifier} for controllable generation. Compared to the classifier guidance\cite{dhariwal2021diffusion}, both training and sampling processes are simplified as it eliminates the need for an external classifier. With the guidance, the conditional data distribution $q(\x| y)$ is learned. In the guided sampling process, the guided noise predictor with guidance label $y$ can be written as 
\begin{equation}
  \bs{\tilde{\epsilon}_{\theta}}(\x_t,t,y) = (1+w)\bs{\epsilon_\theta}(\x_t, t, y) -w\bs{\epsilon_\theta}(\x_t, t, \emptyset),  
\end{equation}
 where $\bs{\epsilon_\theta}(\x_t, t, \emptyset) = \bs{\epsilon_\theta}(\x_t, t)$ indicates no guidance. The parameter $w$ is the guidance strength. The guidance is implemented by utilizing the attention mechanism \cite{vaswani2017attention}.

\section{Methodology}
We demonstrate the guided and unguided forward processes in \cref{sec:forward}. We then introduce the guidance strength selection in \cref{sec:guidance}. Dynamic selections for the noise scale and threshold, along with the aggregation of SAMs, are introduced in  \cref{sec:sams_dynamical}.

\subsection{Guided and unguided forward process}
\label{sec:forward}
 For each step of the forward process, we can obtain the HGP and UGP in one-step mapping as follows:
\begin{align}
  \tilde{\x}_{0,t}^{h} &= \frac{\x_t - \sqrt{1-\bar{\alpha}_{t}}\sbr{(1+w)\bs{\epsilon_\theta}(\x_t, t, h) -w\bs{\epsilon_\theta}(\x_t, t, \emptyset)}}{\sqrt{\bar{\alpha}_{t}}} \label{eq:xh0}\\
    &= \frac{\x_t}{\sqrt{\bar{\alpha}_t}} + B_t\nabla_{\x_t}\log \tilde{p}_\theta\nbr{\x_t| h} \label{eq:xh01}\\
    &= \x_0 + B_t\sbr{(1+w)\nabla_{\x_t}\log p_\theta\nbr{h|\x_t} + \Delta\bs{s}_t} 
  \label{eq:xh02}
\end{align}
\begin{align}
  \tilde{\x}_{0,t}^{\emptyset} &= \frac{\x_t - \sqrt{1-\bar{\alpha}_{t}}\bs{\epsilon_\theta}(\x_t, t, \emptyset)}{\sqrt{\bar{\alpha}_{t}}}\label{eq:xnull0}\\
    &= \frac{\x_t}{\sqrt{\bar{\alpha}_t}}+ B_t\nabla_{\x_t}\log p_\theta\nbr{\x_t},
\end{align}
where $\tilde{p}_\theta\nbr{\x_t| h}\propto p_\theta\nbr{\x_t}p_\theta\nbr{h| \x_t}^{w+1}$, $B_t =  \frac{1-\bar{\alpha}_t}{\sqrt{\bar{\alpha}_t}}$ is a monotonically increasing function. $\nabla_{\x_t}\log p_\theta\nbr{\x_t}= -\frac{1}{\sqrt{1-\bar{\alpha}_t}}\bs{\epsilon_\theta}(\x_t,t,\emptyset)$ according to the Tweedie's Formula \cite{efron2011tweedie}. $\Delta\bs{s}_t =  \nabla_{\x_t}\log p_\theta\nbr{\x_t}- \nabla_{\x_t}\log q\nbr{\x_t}$ is an error term. $p_\theta(h|\x_t)$ represents an implicit classifier induced by the classifier-free guidance \cite{ho2021classifier}. The derivation of \cref{eq:xh01} and \cref{eq:xh02} is detailed in Appendix A. 

We observe that the HGP $\tilde{\x}_{0,t}^{h}$ forces the noised input $\x_t$ to be healthy due to the gradient $\nabla_{\x_t}\log \tilde{p}_\theta\nbr{\x_t| h}$, which is equivalent to shifting the input $\x_0$ towards to its healthy counterpart, in one-step mapping. The gradient $\nabla_{\x_t}\log p_\theta\nbr{h|\x_t}$ serves as the sensitivity map of the implicit classifier $p_\theta\nbr{h|\x_t}$, highlighting regions of $\x_t$ that significantly impact the classification of health. Initially, the sensitivity map focuses on high-frequency regions that are more susceptible to corruption by noise. As the noise level increases, the high-frequency regions are fully corrupted, and the focus gradually shifts to lower-frequency regions that are corrupted until the image is fully corrupted. This is because the high-frequency regions are more perturbed compared to the low-frequency regions due to the uniform spectral density of Gaussian noise and the gradient $\nabla_{\x_t}\log p_\theta\nbr{h|\x_t}$ shifts the corrupted regions in $\x_t$ towards the healthy distribution. The UGP $\tilde{\x}_{0,t}^{\emptyset}$ is not constrained. Note that \cref{eq:xh0} and \cref{eq:xnull0} are essentially the reversed \cref{eq:ddpmforward} with and without guidance respectively. 

We utilize UGP as a reference to measure the divergence between the HGP and UGP, using the following metric:
\begin{align}
  M_t &= \frac{\left\lVert \tilde{\x}_{0,t}^{h} - \tilde{\x}_{0,t}^{\emptyset} \right\rVert_2^2}{d} \\ 
  &= \frac{B_t^2 (1+w)^2}{d} \left\lVert \nabla_{\x_t}\log p_\theta\nbr{h|\x_t}\right\rVert_2^2,
  \label{eq:metric}
\end{align}
where $d$ is the dimension of the input $\x_0$. The derivation of \cref{eq:metric} can be found in Appendix A. The divergence $M_t$ is essentially the magnitude of weighted gradient of the log-likelihood of the implicit classifier $p_\theta\nbr{h|\x_t}$. In the forward process, we collect $\{M_t\}_{t=1}^{T}$ along with two pixel-level errors, crucial for further parameter tuning, as follows:
\begin{align}
  \bs{e}_t^h &= \tilde{\x}_{0,t}^{h} - \x_0 \\
                &= B_t \nbr{(1+w) \nabla_{\x_t}\log p_\theta\nbr{h|\x_t} + \Delta\bs{s}_t}
  \label{eq:eh}
\end{align}
\begin{align}
  \bs{e}_t^\emptyset &= \tilde{\x}_{0,t}^{\emptyset} - \x_0 \\
                        &= B_t \Delta\bs{s}_t
  \label{eq:ee}
\end{align}
Please refer to Appendix A for a detailed derivation of \cref{eq:eh}.

\subsection{Fixed guidance strength selection} 
\label{sec:guidance}
We first select the fixed guidance strength $w$ by using the classification accuracy between healthy and unhealthy samples. This involves calculating the cosine similarity between sequences of MSEs for healthy samples $MSE^h_w = \{\lVert \bs{e}^h_t\rVert_2^2/d\}_{t=1}^{T}$ and unhealthy samples $MSE^\emptyset = \{\lVert \bs{e}^\emptyset_t\rVert^2_2/d\}_{t=1}^{T}$ as follows:
\begin{equation}
    Cos\nbr{MSE^h_w, MSE^\emptyset} = \frac{MSE^h_w \cdot MSE^\emptyset}{\|MSE^h_w\|_2 \cdot \|MSE^\emptyset\|_2}.
\end{equation} 
The rationale is that for healthy samples, the cosine similarity should be high, indicating minimal deviation in their error vectors compared to unhealthy samples. The gradient $\nabla_{\x_t}\log p_\theta\nbr{h|\x_t}$ should exhibit lower sensitivity in healthy samples. If the guidance strength $w$ is too large, the error $\bs{e}_t^h$ will increase, resulting in lower similarity between $MSE^h_w$ and $MSE^\emptyset$ for healthy samples, thus complicating the distinction between healthy and unhealthy states. However, the larger guidance enhances the detection of anomalous regions.

We determine the optimal guidance strength $w^*$ through grid search. On the validation set, we compute the cosine similarity for each sample for each candidate guidance strength $w$ to establish thresholds $Cos_{w}$ that yields best classification accuracy. On the testing set without image-level labels, we use the threshold $Cos_{w^*}$ corresponding to the selected guidance strength $w^*$ to classify inputs. Only samples classified as unhealthy undergo further segmentation, while those predicted as healthy are assigned all-zero masks for performance evaluation purposes. A higher guidance strength $w$ typically results in better segmentation performance for unhealthy samples but may lead to poorer performance when considering both healthy and unhealthy samples due to misclassification. Therefore, our selection strategy aims to strike a balance by choosing the largest guidance strength that does not significantly reduce accuracy compared to the best accuracy obtained with a smaller candidate. This search typically starts with a small guidance strength, which is incrementally increased to find the optimal value $w^*$. The selection process is detailed in Algorithm 1, as presented in Appendix C.

% $Cos\nbr{MSE^h, MSE^\emptyset}$ to appropriately set the guidance strength.

\subsection{Dynamical hyperparameter selection }
\label{sec:sams_dynamical}
The whole dynamical selection process is demonstrated in \cref{fig:diagram}. On the testing set, each sample predicted as unhealthy undergoes both the HFP and UFP to determine the maximal divergence $M_{t_e}$ for hyperparameter selection. To enhance the signal strength and detection accuracy of anomalous regions, we aggregate SAMs. These SAMs are computed at each step of the HFP as the square of the pixel-level errors between the HGP and original inputs, and can be further derived as follows:
\begin{equation}
  \bs{S}_t = \nbr{\bs{e}_t^h}^2 = B_t^2 \nbr{(1+w) \nabla_{\x_t}\log p_\theta\nbr{h|\x_t} + \Delta\bs{s}_t}^2.
  \label{eq:sam}
\end{equation}
As $t$ increases, the sensitive map will focus on lower-frequency regions that are more likely to be anomalous due to changes in texture that are typical of anomalous regions. In each SAM, the signals from healthy regions that are mistakenly targeted by the implicit classifier typically appear randomly distributed. In contrast, signals from anomalous regions exhibit more consistency, indicating that anomalous regions are correctly detected by the implicit classifier. This consistency is crucial to the effectiveness of the aggregation process.

We then aggregate these SAMs to obtain the AAM 
\begin{equation}
  \bs{H} = \frac{1}{t_e}\sum_{t=1}^{t_e} \bs{S}_t,
\end{equation} 
where $t_e$ is the end step of the aggregation, i.e., noise scale. The noise scale here controls the signal strength of the anomalous regions in the AAM. We hope that the SAM $\bs{S}_{t_e}$ can be a `sweet spot' that balances the signal strength of anomalous regions and mistakenly targeted healthy regions. Then, the signal strength of randomly distributed healthy regions can be reduced by the aggregation, while the signal strength of the anomalous regions is maximized because of consistency.

We select the end step $t_e$ as
\begin{equation}
    t_e = \arg\max_{t} M_t.
\end{equation}
We observe that $B_t(1+w)\nabla_{\x_t}\log p_\theta\nbr{h|\x_t}$ is the term that directly contributes to the divergence between $\tilde{\x}_{0,t}^{h}$ and $\x_0$ in \cref{eq:eh}. The largest value $M_{t_e}$ tends to yield the largest changes between $\tilde{\x}_{0,t}^{h}$ and the original input $\x_0$ in terms of being healthy.

To segment the anomalous regions from the AAM, we need to find a threshold $Q^*$ such that the predicted pixel-level anomalous labels is obtained as $\bs{H} \geq Q^*$. We note that the size of the anomalous regions in the input $\x_0$ is roughly proportional to the maximum value of $M_{t_e}$. Hence, the value $M_{t_e}$ serves as a rough indication of the size of the anomalous regions. A smaller quantile of $\bs{H}$ is selected for a larger anomalous region to include more possible pixels since the signal strength of anomalous regions may not be consistent in $\bs{H}$, e.g., the signal strength of central anomalous regions is stronger than the edge due to the proposed aggregation. We select the segmentation threshold $Q^*$ for each predicted unhealthy input $\x_0$ by \cref{alg:quan}. The input $M_{max}$, the largest $M_{t_e}$ of all validation samples, is obtained from the validation set for scaling and is fixed for the testing set. In our case, we set $a=0.90$, $b=0.98$ for the range of quantile. The whole segmentation process is detailed in Algorithm 2, as presented in Appendix C.
\RestyleAlgo{boxruled}
\begin{algorithm}[ht]
  \caption{Selection of quantile $Q$ for a single input $\x_0$}
  \textbf{Input:} $M_{max}$, $a$, $b$, $\bs{H}$\\
  $range = reverse(linspace(a, b, 101))$ \# Set quantile range \\
  $M_s = clamp\nbr{\frac{M_{t_e}}{M_{max}}, 0, 1}$ \\
  $index = round(M_s, 2) \times 100$ \# Keep 2 digits\\
  \textbf{Return} $Q^* = quantile(\bs{H}, range\sbr{index})$
  \label{alg:quan}
\end{algorithm}

\section{Experiments}

\subsection{Dataset and preprocessing}
We evaluated our methods on both BraTS21 dataset \cite{bakas2018identifying} and  ATLAS v2.0 dataset \cite{liew2022large}. BraTS21 dataset are three-dimensional Magnetic Resonance (MR) brain images depicting subjects afflicted with a cerebral tumor. Each subject undergoes scanning through four distinct MR sequences, specifically T1-weighted, T2-weighted, FLAIR, and T1-weighted with contrast enhancement. Given our emphasis on a two-dimensional methodology, our analysis is confined to axial slices. There are 1,254 patients and we split the dataset into 939 patients for training, 63 patients for validation, 252 patients for testing. We randomly select 1,000 samples in validation set for hyperparameter tuning and 10,000 samples in testing set for evaluation. For training, we stack all four modalities while only FLAIR and T2-weighted modalities are used in inference. The ATLAS v2.0 dataset consists of three-dimensional, T1-weighted MR brain images of stroke cases, which present a more challenging scenario for anomaly detection due to data quality and the characteristics of stroke lesions. The dataset contains 655 subjects with manually-segmented lesion mask. We split the dataset into 492 subjects for training, 33 subjects for validation, 130 subjects for testing. We randomly select 1,000 samples in validation set for hyperparameter tuning and 4,000 samples in testing set for evaluation. The data preprocessing and training setup is demonstrated in Appendix B.

\subsection{Dynamical noise scale captures the `sweet spot'}
\begin{figure*}[!h]
  \centering
  \includegraphics[scale=0.152]{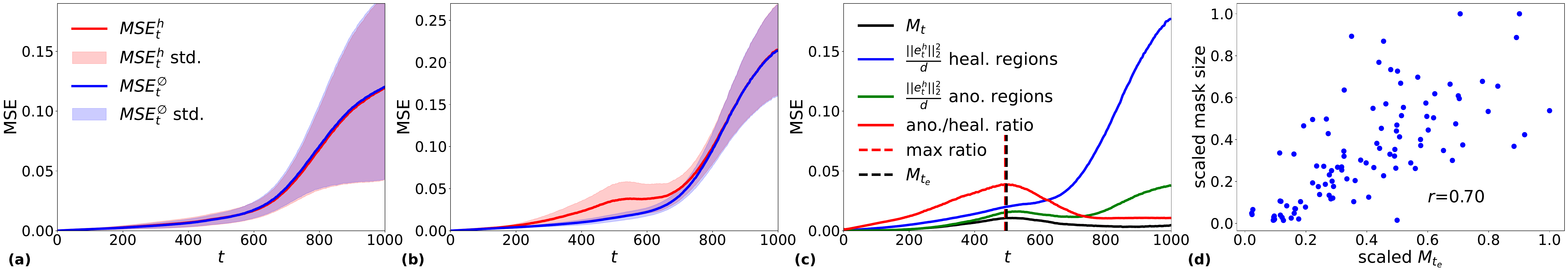}
  \caption{$MSE^h_t$ and $MSE_t^\emptyset$ for (a) healthy samples and (b) unhealthy samples. (c) Average change in the magnitude of SAMs in healthy and anomalous regions, alongside their ratio and divergence $M_t$ between HGP and UGP. (d) Relationship between the size of the anomalous region and the maximal divergence $M_{t_e}$.}
  \label{fig:ablation}
\end{figure*}

Here, we demonstrate that our selected noise scale $t_e$ effectively captures the 'sweet spot' that balances the signal strength between the anomalous and healthy regions in the SAM at $t_e$. To validate this, we analyze 200 unhealthy samples from the BraTS21 dataset. We present the average change of divergence $M_t$ between HGP and UGP, along with the magnitude of SAMs, calculated as $\frac{\lVert\bs{e}_t^h\rVert_2^2}{d}$, in healthy and anomalous regions separately. Additionally, we include the ratio of these magnitude to further illustrate the balance achieved. These results are depicted in \cref{fig:ablation}c. The maximal ratio is almost achieved at $t_e$, indicating that the noise scale $t_e$ effectively captures the 'sweet spot'. We also show the relation between the size of the anomalous region and the maximal divergence $M_{t_e}$ in \cref{fig:ablation}d. We observe that the size of the anomalous region is roughly proportional to the maximal divergence $M_{t_e}$ and we achieved Pearson correlation $r=0.7$. 

\begin{figure*}[!h]
  \centering
  \begin{subfigure}{0.49\linewidth}
    \begin{tikzpicture}
      % Include the main image
      \node at (-1.5,0) {\includegraphics[width=1\textwidth]{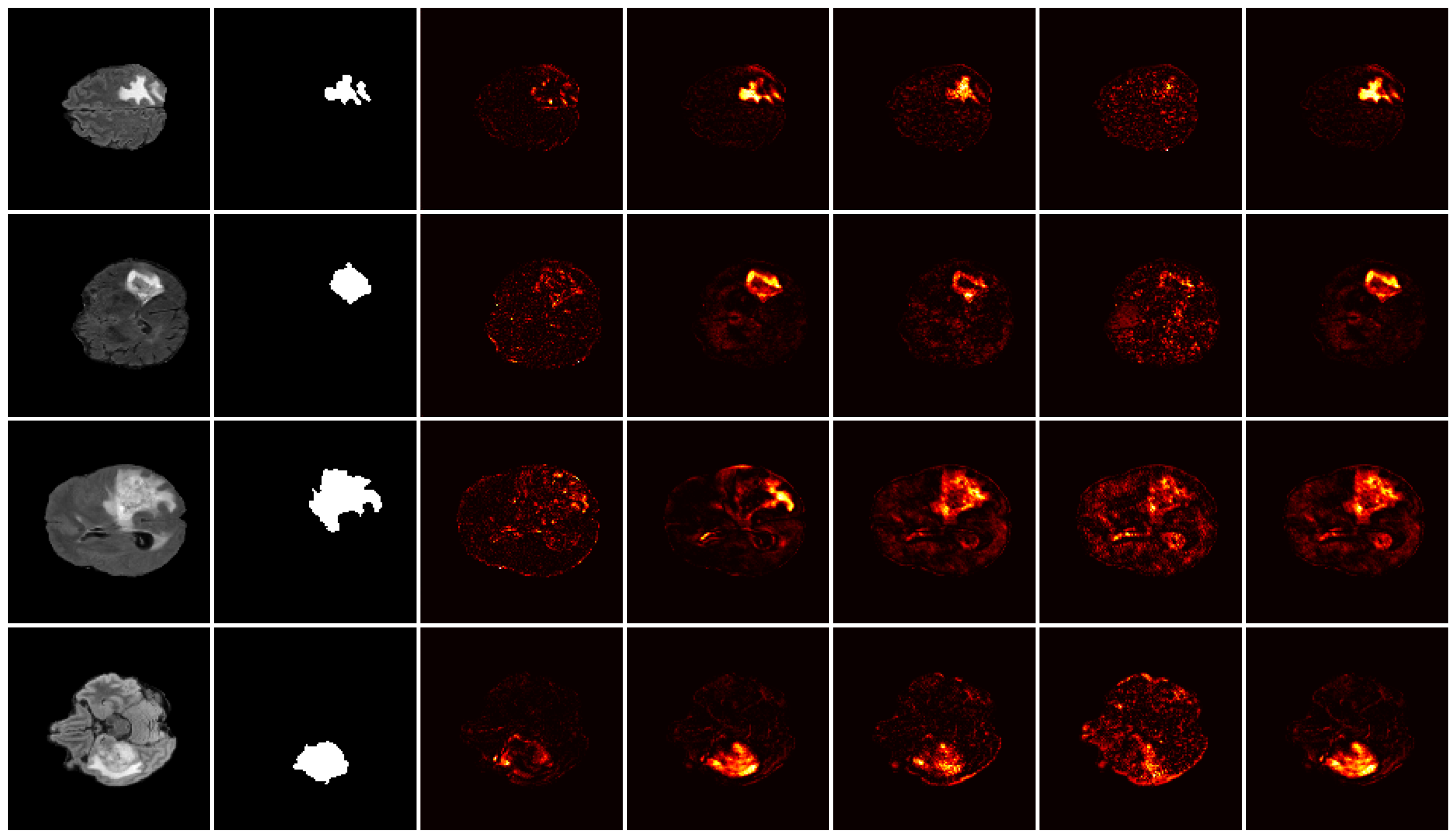}};
      % Adding labels on the top side
      \node at (-5.1,2.60) {\small Input};
      \node at (-3.9,2.60) {\small GT};
      \node at (-2.7,2.60) {\small $\bs{S}_{300}$};
      \node at (-1.5,2.60) {\small $\bs{S}_{500}$};
      \node at (-0.3,2.60) {\small $\bs{S}_{600}$};
      \node at (1.0,2.60) {\small $\bs{S}_{700}$};
      \node at (2.2,2.60) {\small $\bs{S}_{t_e}$};
    \end{tikzpicture}
    \caption{The first two columns show the unhealthy inputs and the ground truth masks. The third to the last columns show the corresponding SAMs at the step $t=100,300,600,700,T$ and $t_e$.}
    \label{fig:SAMs}
  \end{subfigure}
  \hfil
  \begin{subfigure}{0.49\linewidth}
    \begin{tikzpicture}
      % Include the main image
      \node at (-1.5,0) {\includegraphics[width=1\textwidth]{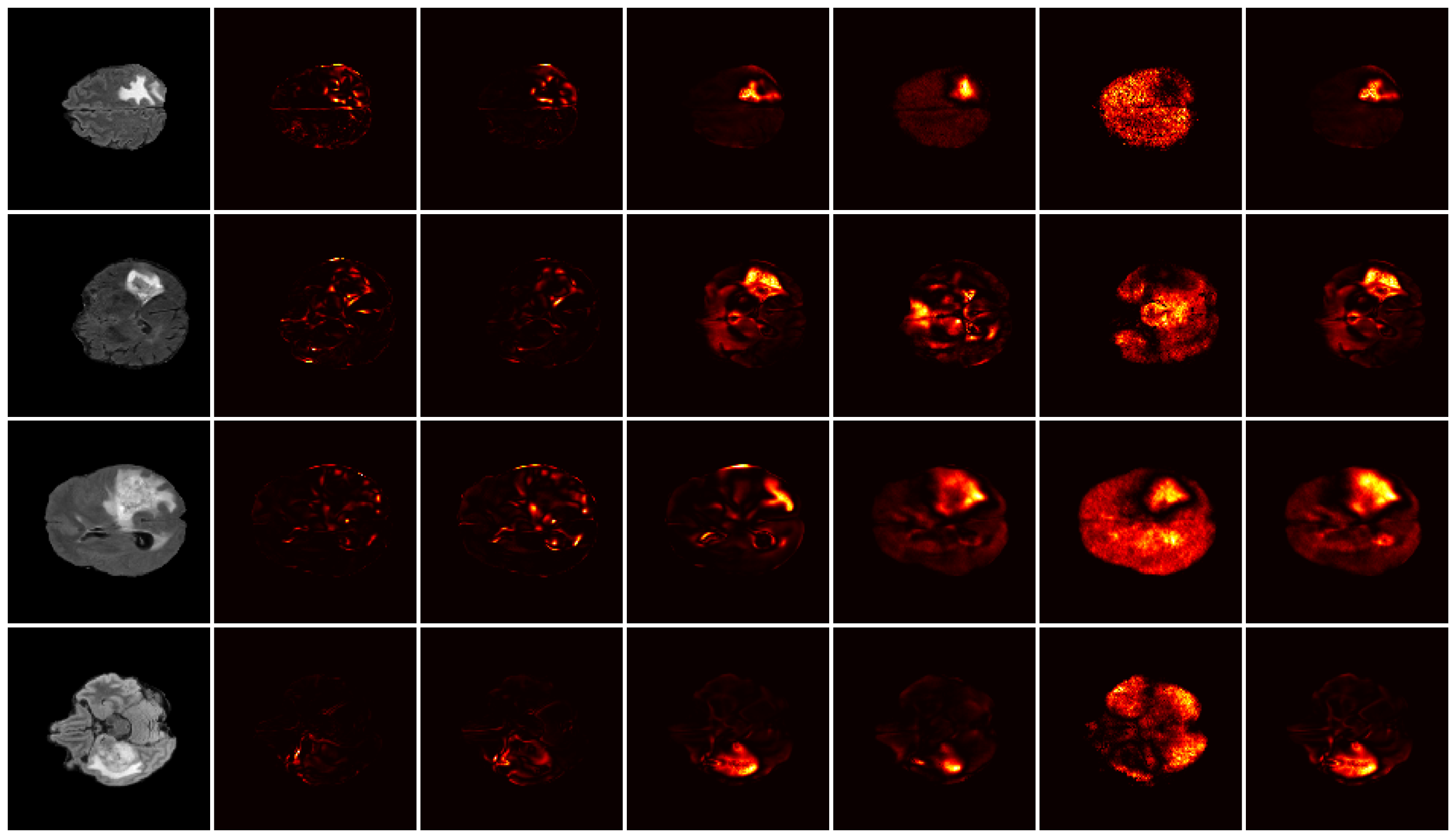}};
      % Adding labels on the top side
      \node at (-5.1,2.60) {\small Input};
      \node at (-3.9,2.60) {\small $\nabla_{\x_{200}}$};
      \node at (-2.7,2.60) {\small $\nabla_{\x_{300}}$};
      \node at (-1.5,2.60) {\small $\nabla_{\x_{500}}$};
      \node at (-0.3,2.60) {\small $\nabla_{\x_{600}}$};
      \node at (1.0,2.60) {\small $\nabla_{\x_{700}}$};
      \node at (2.2,2.60) {\small $\nabla_{\x_{t_e}}$};
    \end{tikzpicture}
    \caption{The first column shows the unhealthy inputs. The third to the last columns show the corresponding gradients $\nabla_{\x_t}\log p_\theta(h|\x_t)|_{t=t_e}$ at the step $t=200,300,500,600,700$ and $t_e$.}
    \label{fig:gradient}
  \end{subfigure}
  \caption{The gradient of the log-likelihood of the implicit classifier (a) and SAMs (b) at the selected steps from BraTS21 dataset}
\end{figure*}

We display samples of SAMs alongside the corresponding gradient of the log-likelihood of the implicit classifier $\nabla_{\x_t}\log p_\theta(h|\x_t)$ at selected steps from BraTS21 dataset in \cref{fig:SAMs} and \cref{fig:gradient}. We observe that the implicit classifier predominantly targets anomalous regions at $t=500$ and $t=600$. At $t=200$ and $t=300$, the gradient focuses on high-frequency regions, transitioning to significantly lower-frequency regions at $t=700$. As noted earlier, the anomalous regions exhibit consistent distribution patterns, while the healthy regions appear randomly distributed across different frequency levels. The SAM and its corresponding gradient at step $t_e$ strikes a balance between the signal strength of healthy regions and anomalous regions. However, the extra error terms $\Delta\bs{s}_t$ in the SAMs may correct the gradient term and this effect is obvious at $t=700$.

\subsection{Main results}
We assess our model’s performance using pixel-level metrics, including the DICE score, intersection over union (IoU), and area under the precision-recall curve (AUPRC) with a specific focus on the foreground areas. Our evaluation on the BraTS21 test set, which comprises 10,000 samples, are comprehensively detailed in two separate configurations: the mixed setup, which includes all samples, and the unhealthy setup, which exclusively considers unhealthy samples. These results are presented in \cref{tab:results_bra}. Similarly, for the ATLAS dataset, which comprises 4,000 mixed samples including 1,003 unhealthy and 2,997 healthy samples, we present the performance results separately for the mixed setup and the unhealthy setup in \ref{tab:results_atlas}.

For the proposed method, we present results for four setups: (i) \textbf{AnoFPDM (DDPM)} (stochastic encoding): \cref{eq:ddpmforward} is used to perturb inputs; (ii) \textbf{AnoFPDM (DDIM)} (deterministic encoding): the inputs are noised by \cref{eq:ddimforward}; (iii) \textbf{AnoFPDM (DDIM $\bs{S}_{t_e}$)}: similar to \textbf{AnoFPDM (DDIM)}, but segmentation is performed using only the SAM at $t_e$; (iv) \textbf{AnoFPDM (DDPM $\bs{S}_{t_e}$)}: follows the \textbf{AnoFPDM (DDPM)} setup, with segmentation using only the SAM at $t_e$. For comparative analysis, we report the results from AnoDDPM \cite{wyatt2022anoddpm} with simplex noise and Gaussian noise for fair comparison, DDIM with classifier guidance \cite{wolleb2022diffusion}, DDIM with classifier-free guidance \cite{sanchez2022healthy} and classifier guided conditional diffusion model (CG-CDM) \cite{hu2023conditional}. The first two methods are only trained on healthy samples. The hyperparameters for all comparison methods are optimized using a grid search on 1,000 mixed samples with pixel-level labels for both setups. Notably, our methods do not require pixel-level labels for tuning. The postprocessing for segmentation is detailed in Appendix F.

Our method, especially AnoFPDM (DDIM) with deterministic encoding, demonstrates superior performance in terms of DICE and IoU scores, particularly in scenarios only involving unhealthy samples. AnoDDPM with simplex noise achieves AUPRC scores comparable to ours. However, its segmentation performance in unhealthy settings is hindered by a fixed threshold. Across both datasets, our approach consistently balances performance between mixed and unhealthy setups, notably excelling in the ATLAS v2.0 dataset where other comparison methods struggle to detect anomalous regions. We discuss further variations in performance across both datasets in Appendix G. The qualitative results, displayed in \cref{fig:qualitative}, underscore our method's ability to enhance the signal strength of anomalous regions, which is confirmed by higher AUPRC scores. Notably, the signal strength at the edges of anomalous regions is weaker compared to their central parts. This effect becomes more pronounced in larger anomalous regions. More qualitative results are exhibited in Appendix E.
\begin{table*}[!h]
  \small{
  \begin{subtable}[t]{\linewidth}
    \centering
    \begin{tabular}{l l c c c c c c}
       \toprule
       &&  \multicolumn{3}{c}{Mixed} & \multicolumn{3}{c}{Unhealthy}\\
       \cmidrule(lr){3-5} \cmidrule(lr){6-8}
       \textbf{Type} & \textbf{Methods} & DICE & IoU & AUPRC & DICE & IoU & AUPRC\\
       \midrule
       \multirow{4}{*}{Reconstruction} & \xmark\space AnoDDPM (G) \cite{wyatt2022anoddpm} & 66.1$\pm0.1$ & 61.7$\pm0.1$ & 51.8$\pm0.1$ & 37.6$\pm$0.1 & 28.1$\pm$0.1 & 61.3$\pm$0.1\\
       & \xmark\space AnoDDPM (S) \cite{wyatt2022anoddpm} & 75.1$\pm$0.3 & 69.5$\pm0.2$ & 67.3$\pm$0.1 & 53.7$\pm$2.7 & 45.5$\pm$1.3 & 71.8$\pm$0.1\\
       & \xmark\space DDIM clf \cite{wolleb2022diffusion} & \underline{76.5$\pm$0.1} & \underline{71.0$\pm$0.1} & 58.4$\pm$0.3 & 52.2$\pm$0.2 & 40.4$\pm$0.2 & 61.6$\pm$0.2\\
       & \xmark\space DDIM clf-free \cite{sanchez2022healthy} & 74.3$\pm$0.0 & 69.1$\pm$0.0 & 59.9$\pm$0.0 & 49.1$\pm$0.0 & 38.1$\pm$0.0 & 61.4$\pm$0.0\\
       \midrule
       \multirow{5}{*}{Gradient} & \xmark\space CG-CDM \cite{hu2023conditional} & - & - & - &44.4$\pm$0.3 &32.2$\pm$0.5 &31.2$\pm$0.7\\
      & \cmark\textbf{AnoFPDM (DDPM)}  &75.2$\pm$0.3& 68.5$\pm$0.3& 68.3$\pm$0.1 & 57.5$\pm$0.4& 46.0$\pm$0.2 & \underline{72.9$\pm$0.1}\\
      & \cmark\textbf{AnoFPDM (DDPM $\bs{S}_{t_e}$)} &70.1$\pm$0.1 &64.6$\pm$0.3 & 56.7$\pm$0.4& 47.8$\pm$0.1 & 36.0$\pm$0.1 &61.1$\pm$0.5\\ 
      & \cmark\textbf{AnoFPDM (DDIM)} & \textbf{77.4$\pm$0.0} & \textbf{72.5$\pm$0.0} &  \textbf{72.2$\pm$0.0}& \textbf{61.5$\pm$0.0}& \textbf{50.0$\pm$0.1} & \textbf{75.5$\pm$0.0}\\
      & \cmark\textbf{AnoFPDM (DDIM $\bs{S}_{t_e}$)} &  75.7$\pm$0.0 & 70.6$\pm$0.0 & \underline{69.9$\pm$0.1} & \underline{58.7$\pm$0.0}& \underline{47.0$\pm$0.0}& 72.6$\pm$0.1\\
      \bottomrule
    \end{tabular}
    \caption{Segmentation performance on mixed and unhealthy samples from BraTS21 dataset.}
    \label{tab:results_bra}
\end{subtable}

\begin{subtable}[c]{\linewidth}
  \centering
  \begin{tabular}{l l c c c c c c}
      \toprule
      && \multicolumn{3}{c}{Mixed} & \multicolumn{3}{c}{Unhealthy}\\
      \cmidrule(lr){3-5} \cmidrule(lr){6-8}
      \textbf{Type} & \textbf{Methods} & DICE & IoU & AUPRC & DICE & IoU & AUPRC\\
      \midrule
      \multirow{4}{*}{Reconstruction} & \xmark\space AnoDDPM (G) \cite{wyatt2022anoddpm} &74.8$\pm$0.1 &\underline{74.8$\pm$0.1} &2.0$\pm$0.1 &0.4$\pm$0.1 &0.2$\pm$0.1&6.5$\pm$0.2  \\
      & \xmark\space AnoDDPM (S) \cite{wyatt2022anoddpm} &\underline{74.9$\pm$0.1}  &74.6$\pm$0.1  &20.8$\pm$0.5  &3.4$\pm$1.0  &3.3$\pm$0.7 &\underline{30.9$\pm$0.4} \\
      & \xmark\space DDIM clf \cite{wolleb2022diffusion} &51.5$\pm$0.8  &50.8$\pm$0.7  &1.9$\pm$0.1  &5.8$\pm$0.1  &3.7$\pm$0.1 &5.6$\pm$0.1 \\
      & \xmark\space DDIM clf-free \cite{sanchez2022healthy}  &73.5$\pm$0.0 &73.0$\pm$0.0 &9.3$\pm$0.0 &0.1$\pm$0.0 &0.1$\pm$0.0 &13.6$\pm$0.0\\
      \midrule
      \multirow{5}{*}{Gradient} & \xmark\space CG-CDM \cite{hu2023conditional} & - & - & - &2.1$\pm$0.0 &1.1$\pm$0.0 &1.6$\pm$0.0\\
      & \cmark\textbf{AnoFPDM (DDPM)} &74.5$\pm$0.1 &74.5$\pm$0.1 &\underline{22.4$\pm$0.1} &\underline{17.5$\pm$0.1} &\underline{12.4$\pm$0.1} &29.2$\pm$0.1 \\
      & \cmark\textbf{AnoFPDM (DDPM $\bs{S}_{t_e}$)} &74.6$\pm$0.1 &74.6$\pm$0.1 &6.1$\pm$0.1 &9.5$\pm$0.1 &6.4$\pm$0.1 &10.8$\pm$0.2\\ 
      & \cmark\textbf{AnoFPDM (DDIM)} &\textbf{75.5$\pm$0.2} &\textbf{75.5$\pm$0.2} &\textbf{22.5$\pm$0.1} &\textbf{21.5$\pm$0.0} &\textbf{15.5$\pm$0.0} &\textbf{31.2$\pm$0.1} \\
      & \cmark\textbf{AnoFPDM (DDIM $\bs{S}_{t_e}$)} &74.5$\pm$0.2 &74.5$\pm$0.2 &2.0$\pm$0.1 &9.8$\pm$0.1 &6.9$\pm$0.1 &4.1$\pm$0.1 \\  
      \hline
    \end{tabular}
    \caption{Segmentation performance on mixed slices and unhealthy samples from ATLAS v2.0 dataset.}
    \label{tab:results_atlas} 
\end{subtable}
  \caption{Quantitative results across both datasets. Standard deviations are derive from three runs. The best and second-best performances are in bold and underline, respectively. Methods requiring pixel-level labels for tuning are marked with \cmark, those that do not with \xmark. CG-CDM \cite{hu2023conditional} is tested only on the unhealthy samples for consistency with its original study.}
  }
\end{table*}

\begin{figure*}[!h]
  \centering
  \begin{subfigure}{0.49\linewidth}
    \begin{tikzpicture}
      % Include the main image
      \node at (-1.4,0.3) {\includegraphics[width=1\textwidth]{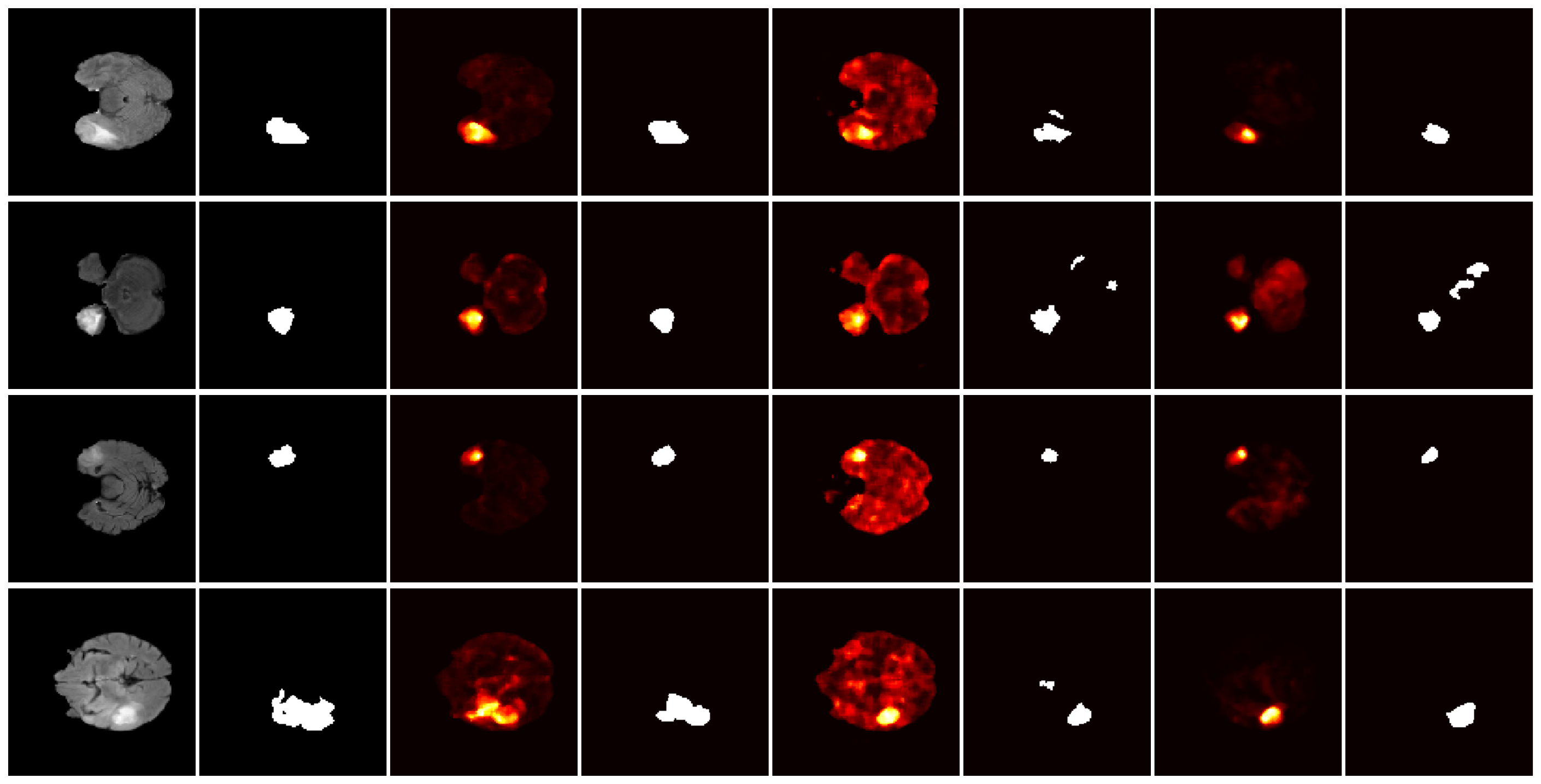}};
      % Adding labels on the top side
      \node at (-5.1,2.60) {\small Input};
      \node at (-4.05,2.60) {\small GT};
      \node at (-2.5,2.78) {
        \small 
        \begin{minipage}{2cm}
                  \centering
                  Ours\\
                  (DDIM)
        \end{minipage}};
      \node at (-0.4,2.78) {
        \small 
        \begin{minipage}{2cm}
                  \centering
                  DDIM\\
                  clf-free
        \end{minipage}};
      \node at (1.8,2.78) {
        \small 
        \begin{minipage}{2cm}
                  \centering
                  Ano\\
                  DDPM (S)
        \end{minipage}};
    \end{tikzpicture}
    \label{fig:qualitative_brats}
  \end{subfigure}
  \hfil
  \begin{subfigure}{0.49\linewidth}
    \begin{tikzpicture}
      % Include the main image
      \node at (-1.4,0.3) {\includegraphics[width=1\textwidth]{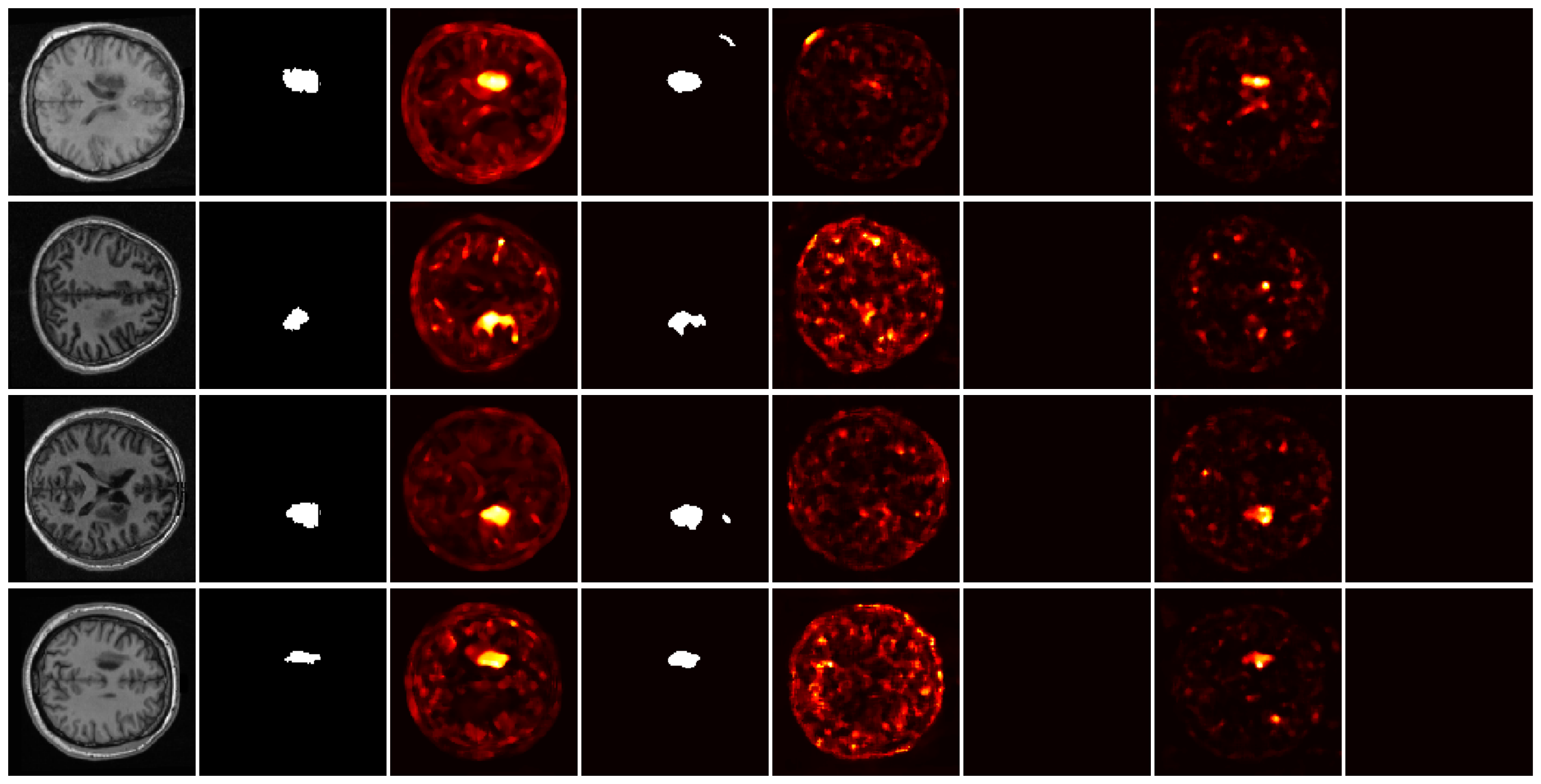}};
      % Adding labels on the top side
      \node at (-5.1,2.60) {\small Input};
      \node at (-4.05,2.60) {\small GT};
      \node at (-2.5,2.78) {
        \small 
        \begin{minipage}{2cm}
                  \centering
                  Ours\\
                  (DDIM)
        \end{minipage}};
      \node at (-0.4,2.78) {
        \small 
        \begin{minipage}{2cm}
                  \centering
                  DDIM\\
                  clf-free
        \end{minipage}};
      \node at (1.8,2.78) {
        \small 
        \begin{minipage}{2cm}
                  \centering
                  Ano\\
                  DDPM (S)
        \end{minipage}};
    \end{tikzpicture}
    \label{fig:qualitative_atlas}
  \end{subfigure}
  \caption{Qualitative Comparison of Anomaly Maps and Segmentation. (a) From the BraTS21 dataset and (b) from the ATLAS v2.0 dataset. The first column displays the original input images, and the second column shows the corresponding ground truth for anomaly segmentation. Subsequent columns present the anomaly maps and segmentation results obtained using our method, AnoFPDM with the DDIM setting, alongside those from the second and third best comparative methods. Each row represents a different sample.}
  \label{fig:qualitative}
\end{figure*}

\subsection{Ablation study}
\begin{figure}
  \centering
  \includegraphics[scale=0.133]{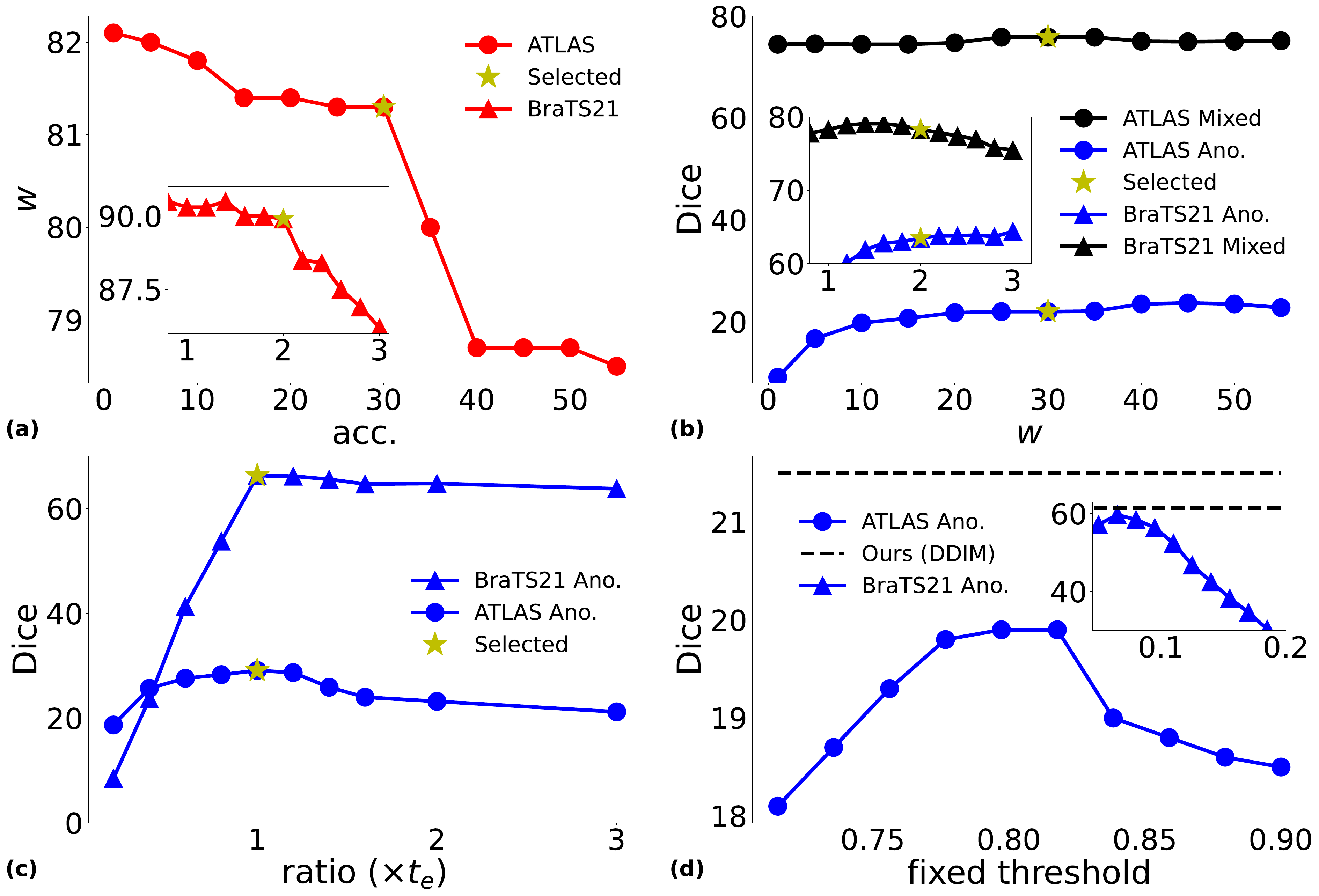}
  \caption{Analysis of the impact of hyperparameter variations on model performance. (a) Variation in classification accuracy as a function of guidance strength $w$. (b) Corresponding changes in DICE score with different guidance stength $w$. (c) Effect of the end step $t_e$ on DICE score, using the selected guidance strength. (d) Comparison of DICE scores using a fixed threshold.}
  \label{fig:ablation_2}
\end{figure}

To validate the effectiveness of our hyperparameter selection, we conducted a series of performance comparisons using different guidance strength $w$, end step $t_e$, and threshold $Q^*$ across both datasets. The results are displayed in \cref{fig:ablation_2}. Specifically, we first examined the impact of guidance strength $w$ in \cref{fig:ablation_2}a and \cref{fig:ablation_2}b. For the BraTS21 and ATLAS v2.0 datasets, we selected the guidance strength $w=2$ and $w=30$ respectively. These values represent the maximum at which the accuracy, based on cosine similarity, does not significantly decline, staying within 99\% of the peak accuracy. Additionally, we assessed the MSEs $MSE_t^h$ and $MSE_t^\emptyset$ at $w=2$ for both healthy and unhealthy samples from BraTS21 dataset in \cref{fig:ablation}a and \cref{fig:ablation}b. These figures highlight noticeable differences in the unhealthy samples, whereas the differences in the healthy samples are less pronounced. We further explored the variation in the DICE score across different guidance strengths in \cref{fig:ablation_2}b, with other hyperparameters being dynamically adjusted per our proposed method. The DICE score varies across the two setups, namely the mixed and unhealthy setups, with different guidance strengths $w$, confirming our statement that a higher guidance strength enhances the detection of anomalous regions. However, the overall performance in the ATLAS v2.0 dataset is less impacted by changes in guidance strength compared to the BraTS21 dataset, which can be attributed to their differing distributions of healthy and unhealthy samples. Our selection tends to strikes a balance between the two setups.

The impact of the end step $t_e$ on the DICE score for both datasets with the selected guidance strength is depicted in \cref{fig:ablation_2}c. The maximal DICE score is achieved at $t_e$, followed by a slight decline as $t_e$ extends to $3\times t_e$ because more noise is introduced into the AAM. This pattern validates the selection of $t_e$, confirming it effectively enhances the signal strength of anomalous regions in the AAM. We also show the influence of the threshold on the DICE score in \cref{fig:ablation_2}d. We compare our dynamical selection $Q^*$ with the fixed threshold selection. Given the aggregated anomaly map obtained by the proposed method, we calculate the DICE score with various fixed thresholds. Our dynamic threshold selection significantly outperforms the fixed threshold approach, thereby validating the efficacy of our method in optimizing threshold selection.

We also discuss the use of the gradient directly as SAMs for segmentation in Appendix D.

\section{Conclusion}
We propose a novel weakly-supervised method, AnoFPDM, for anomaly detection. This method dynamically selects the noise scale and threshold for each input while maintaining a fixed guidance strength selection. In quantitative evaluations, our method surpasses previous approaches and demonstrates enhanced signal strength of anomalous regions in qualitative assessments. Unlike traditional methods, our selection process does not involve pixel-level labels, making it more practical for real-world applications. We also discuss the limitations in Appendix G.

%%%%%%%%% REFERENCES
{\small
\bibliographystyle{ieee_fullname}
\bibliography{egbib}
}

%%%%%%%%% APPENDIX
\clearpage
\appendix
\onecolumn
\section*{\huge{Appendices}} 

\section{Derivation}
We provide the derivation of Eq. (8), (9), (13) and (15) in the main text.
For Eq. (8), we have
\begin{align}
   \tilde{\x}_{0,t}^{h} &= \frac{\x_t - \sqrt{1-\bar{\alpha}_{t}}\sbr{(1+w)\bs{\epsilon_\theta}(\x_t, t, h) -w\bs{\epsilon_\theta}(\x_t, t, \emptyset)}}{\sqrt{\bar{\alpha}_{t}}}\nonumber\\
     &= \frac{\x_t}{\sqrt{\bar{\alpha}_t}} - A_t\sbr{(1+w)\bs{\epsilon_\theta}(\x_t, t, h) -w\bs{\epsilon_\theta}(\x_t, t, \emptyset)} \nonumber\\ 
     &= \frac{\x_t}{\sqrt{\bar{\alpha}_t}} + B_t\sbr{(1+w)\bs{s_\theta}(\x_t, t, h) - w\bs{s_\theta}(\x_t, t, \emptyset)}\nonumber\\
     &= \frac{\x_t}{\sqrt{\bar{\alpha}_t}} + B_t\sbr{(1+w)\nabla_{\x_t}\log p_\theta\nbr{\x_t\mid h} - w\nabla_{\x_t}\log p_\theta\nbr{\x_t}}\nonumber\\
     &= \frac{\x_t}{\sqrt{\bar{\alpha}_t}} + B_t\sbr{(1+w)\nabla_{\x_t}\log p_\theta\nbr{h\mid\x_t} + \nabla_{\x_t}\log p_\theta\nbr{\x_t}}\nonumber\\
     &= \frac{\x_t}{\sqrt{\bar{\alpha}_t}} + B_t\nabla_{\x_t}\log \sbr{p_\theta(\x_t)p_\theta\nbr{h\mid\x_t}^{1+w}}\nonumber\\
     &= \frac{\x_t}{\sqrt{\bar{\alpha}_t}} + B_t\nabla_{\x_t}\log \tilde{p}_\theta\nbr{\x_t\mid h}\nonumber
\end{align}
 where $A_t = \frac{\sqrt{1-\bar{\alpha}_t}}{\sqrt{\bar{\alpha}_t}}$ and $B_t = \frac{1-\bar{\alpha}_t}{\sqrt{\bar{\alpha}_t}}$. The score functions are defined as $\bs{s_\theta}(\x_t, t, h) = \nabla_{\x_t}\log p_\theta\nbr{\x_t\mid h}$ and $\bs{s_\theta}(\x_t, t, \emptyset) = \nabla_{\x_t}\log p_\theta\nbr{\x_t}$. For Eq. (9), we substitute $\x_t = \sqrt{\bar{\alpha}_t}\x_0 + \sqrt{1-\bar{\alpha}_t}\bs{\epsilon}_t$ into Eq. (8).

For Eq. (13) and (15), we have
\begin{align}
   M_t &= \frac{\left\lVert \tilde{\x}_{0,t}^{h} - \tilde{\x}_{0,t}^{\emptyset} \right\rVert_2^2}{d}\nonumber\\ 
   &= \frac{A_t^2 (1+w)^2}{d} \left\lVert \bs{\epsilon_{\theta,t}^h} - \bs{\epsilon_{\theta,t}^\emptyset} \right\rVert_2^2 \nonumber\\
   &= \frac{B_t^2 (1+w)^2}{d} \left\lVert \bs{s^h_{\theta,t}} - \bs{s_{\theta,t}^\emptyset}\right\rVert_2^2 \nonumber\\
   &= \frac{C_t^2}{d} \left\lVert \nabla_{\x_t}\log p_\theta\nbr{\x_t\mid h} - \nabla_{\x_t}\log p_\theta\nbr{\x_t}\right\rVert_2^2\nonumber\\ 
   &= \frac{C_t^2}{d} \left\lVert \nabla_{\x_t}\log p_\theta\nbr{h\mid\x_t}\right\rVert_2^2 \nonumber
\end{align}
where $C_t = B_t (1+w)$
\begin{align}
   \bs{e}_t^h &= \tilde{\x}_{0,t}^{h} - \x_0 \nonumber \\
   &= B_t\sbr{(1+w)\nabla_{\x_t}\log p_\theta\nbr{\x_t\mid h} - w\nabla_{\x_t}\log p_\theta\nbr{\x_t} - \nabla_{\x_t}\log q(\x_t)}\nonumber\\
   &= B_t \nbr{(1+w) \nabla_{\x_t}\log p_\theta\nbr{h\mid\x_t} + \Delta\bs{s}_t} \nonumber
\end{align}

\section{Data Preprocessing and Training Hyperparameters}
We standardized preprocessing for both the BraTS21 and ATLAS v2.0 datasets. Each subject was normalized by dividing it by the 99th percentile intensity of foreground voxels, and pixel values were then scaled to the range of $\sbr{-1, 1}$. All samples are interpolated to $128\times 128$. 

The backbone U-net is adopted from the previous work \cite{sanchez2022healthy}. Our model is trained on 2 Nvidia A100 GPUs with 80GB memory. The training hyperparameters are summarized in Tab. \ref{tab:hyper}, and we used the same hyperparameters for both dataset.
\begin{table}[!h]
    \centering
    \begin{tabular}{ll}
        \toprule
         Diffusion steps&  1000\\
         Noise schedule&   linear\\
         \midrule
         Channels&   128\\
         Heads & 2\\
         Attention resolution&   32,16,8\\
         Channel multiplier& 1, 1, 2, 3, 4\\
         Dropout & 0.1\\
         EMA rate & 0.9999  \\
         \midrule
         Optimiser & AdamW\\
         Learning rate & $1e^{-4}$\\
         $\beta_1$, $\beta_2$ & 0.9, 0.999\\
         \midrule
         Global batch size & 64\\
         Null label ratio & 0.1\\
         \midrule
         dropout & 0.1\\
         \bottomrule
    \end{tabular}
    \caption{Training hyperparameters used in our method.}
    \label{tab:hyper}
\end{table}

\section{Fixed Guidance Selection and Segmentation}
We illustrate the fixed guidance selection in \cref{alg:guidance} and outline the complete segmentation process in \cref{alg:segmentation}.

\RestyleAlgo{boxruled}
\begin{algorithm}[!h]
  \caption{Selection of fixed guidance $w^*$}
  \textbf{Input:} $n$ sorted candidates $[w_1,...,w_n]$, validation set with image-level labels \\
  \textbf{for} each candidate $w_i$:\\
  \quad calculate cosine similarity for each sample in validation set with Eq. 18\\
  \quad classify each samples in validation set with cosine similarity threshold $Cos_{w_i}$\\
  \quad get the maximal classification accuracy $Acc_{w_i}$ using the optimal threshold $Cos^*_{w_i}$\\
  \textbf{end for}\\

  $w = \arg\max_{w} Acc_{w_i}$\\
  $w^* = w_i > w$ with $\frac{Acc_{w_i}}{Acc_{w}}\approx 0.98$\\
  $Cos_{w^*} = Cos^*_{w_i}$ \# corresponding threshold \\
  \textbf{Return} $w^*$, $Cos_{w^*}$
  \label{alg:guidance}
\end{algorithm}

\RestyleAlgo{boxruled}
\begin{algorithm}[!h]
  \caption{The full segmentation process for a single input $\x_0$}
  \textbf{Input:} fixed guidance $w^*$, input $\x_0$\\
  \textbf{for} each time step $t$\\
  \quad $\tilde{\x}_{0,t}^{h} = \frac{\x_t - \sqrt{1-\bar{\alpha}_{t}}\sbr{(1+w^*)\bs{\epsilon_\theta}(\x_t, t, h) -w^*\bs{\epsilon_\theta}(\x_t, t, \emptyset)}}{\sqrt{\bar{\alpha}_{t}}}$ \# Healthy guided prediction Eq. 7\\
  \quad $\tilde{\x}_{0,t}^{\emptyset} = \frac{\x_t - \sqrt{1-\bar{\alpha}_{t}}\bs{\epsilon_\theta}(\x_t, t, \emptyset)}{\sqrt{\bar{\alpha}_{t}}}$ \# unguided prediction Eq. 10\\
  \quad $M_t = \frac{\left\lVert \tilde{\x}_{0,t}^{h} - \tilde{\x}_{0,t}^{\emptyset} \right\rVert_2^2}{d}$ \# divergence Eq. 12\\
  \quad $\bs{S}_t = \tilde{\x}_{0,t}^{\emptyset} - \x_0$ \# obtain SAM using Eq. 19\\
  \textbf{end for}\\

  $t_e = \arg\max_{t} M_t$ \# find the end step\\
  $\bs{H} = \frac{1}{t_e}\sum_{t=1}^{t_e} \bs{S}_t$ \# aggregated SAMs Eq. 20\\
  Obtain the quantile $Q^*$ using Algorithm 1 from the main text\\
  $\text{predicted pixel-level labels} = \bs{H} \geq Q^*$\\
  \textbf{Return} predicted pixel-level labels
  \label{alg:segmentation}
\end{algorithm}

\section{Gradient for Segmentation}
We investigate the gradient $\nabla_{\x_t}\log p_\theta\nbr{h|\x_t}$ as the SAMs for segmentation, while keep other settings unchanged. Here, we use the same BraTS21 testing data, focusing on the unhealthy setup with four configurations: (i) DDIM $\nabla^2\log p_\theta$: the gradient is directly used as SAMs for segmentation; (ii) DDIM $\nabla^2\log p_\theta|_{t={t_e}}$: only the gradient at the end step is ued for segmentation; (iii) DDIM $C_t^2\nabla^2\log p_\theta$: the weighted gradient is used as SAMs; and (iv) DDIM $B_t^2 \nbr{(1+w) \nabla\log p_\theta + \Delta\bs{s}_t}^2$: the original SAMs.

The quantitative results are exhibited in \cref{tab:results_gradient}. We note that the performance of other configurations is significantly lower compared to using the original SAMs. The last setup \textbf{DDIM $C_t\nabla\log p_\theta$}, the weighted gradient, achieved better segmentation results compared to non-weighted gradient SAMs. This is attributed to the weight $C_t$, a monotonically increasing function. The SAMs with anomalous regions are more weighted. Also, the error term $\Delta\bs{s}_t$ used in the original SAMs is not considered here, which may alleviate the false detection by the implicit classifier.
\begin{table}[!h]
   \centering
   {\small{
   \begin{tabular}{l c c c}
       \toprule
       \textbf{Methods} & DICE & IoU & AUPRC\\
       \midrule
       \textbf{DDIM $\nabla^2\log p_\theta$}  &42.7$\pm$0.2 &30.6$\pm$0.1 &42.0$\pm$0.0\\
       \textbf{DDIM $\nabla^2\log p_\theta|_{t={t_e}}$} &53.1$\pm$0.0 &40.4$\pm$0.0&58.8$\pm$0.0\\
       \textbf{DDIM $C_t^2\nabla^2\log p_\theta$} &57.2$\pm$0.1 &45.8$\pm$0.1&70.3$\pm$0.0\\
       $^*$\textbf{DDIM $B_t^2 \nbr{(1+w) \nabla\log p_\theta + \Delta\bs{s}_t}^2$} &\textbf{61.5$\pm$0.0}&\textbf{51.0$\pm$0.1} &\textbf{75.5$\pm$0.1}\\
       \bottomrule
   \end{tabular}}
   \caption{Segmentation performance on unhealthy samples from BraTS21 dataset using the gradient $\nabla_{\x_t}\log p_\theta\nbr{h|\x_t}$ as the SAMs. The last setup with $*$ is the original SAMs.}
   \label{tab:results_gradient}}
 \end{table}

\section{More Qualitative Results}
\begin{figure*}[!h]
   \centering
   \begin{subfigure}{0.9\linewidth}
     \begin{tikzpicture}
       % Include the main image
       \node at (-1.4,0.3) {\includegraphics[width=1\textwidth]{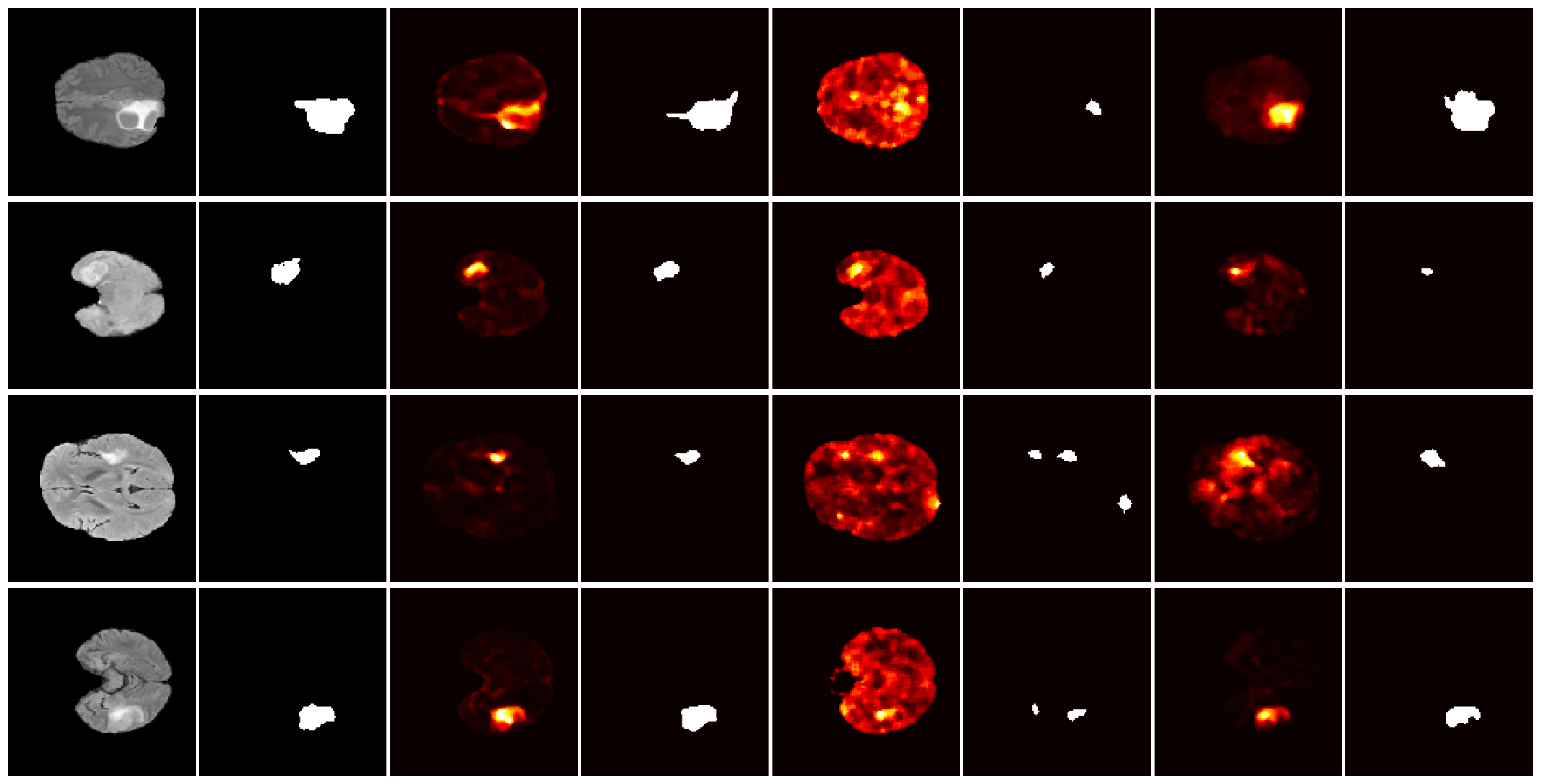}};
       % Adding labels on the top side
       \node at (-8.3,4.40) {Input};
       \node at (-6.30,4.40) {GT};
       \node at (-3.4,4.40) {Ours (DDIM)};
       \node at (0.6,4.40) {DDIM clf-free};
       \node at (4.3,4.40) {Ano DDPM (S)};
     \end{tikzpicture}
     \label{fig:qualitative_brats_app}
   \end{subfigure}
   \hfil
   \begin{subfigure}{0.9\linewidth}
     \begin{tikzpicture}
       % Include the main image
       \node at (-1.4,0.3) {\includegraphics[width=1\textwidth]{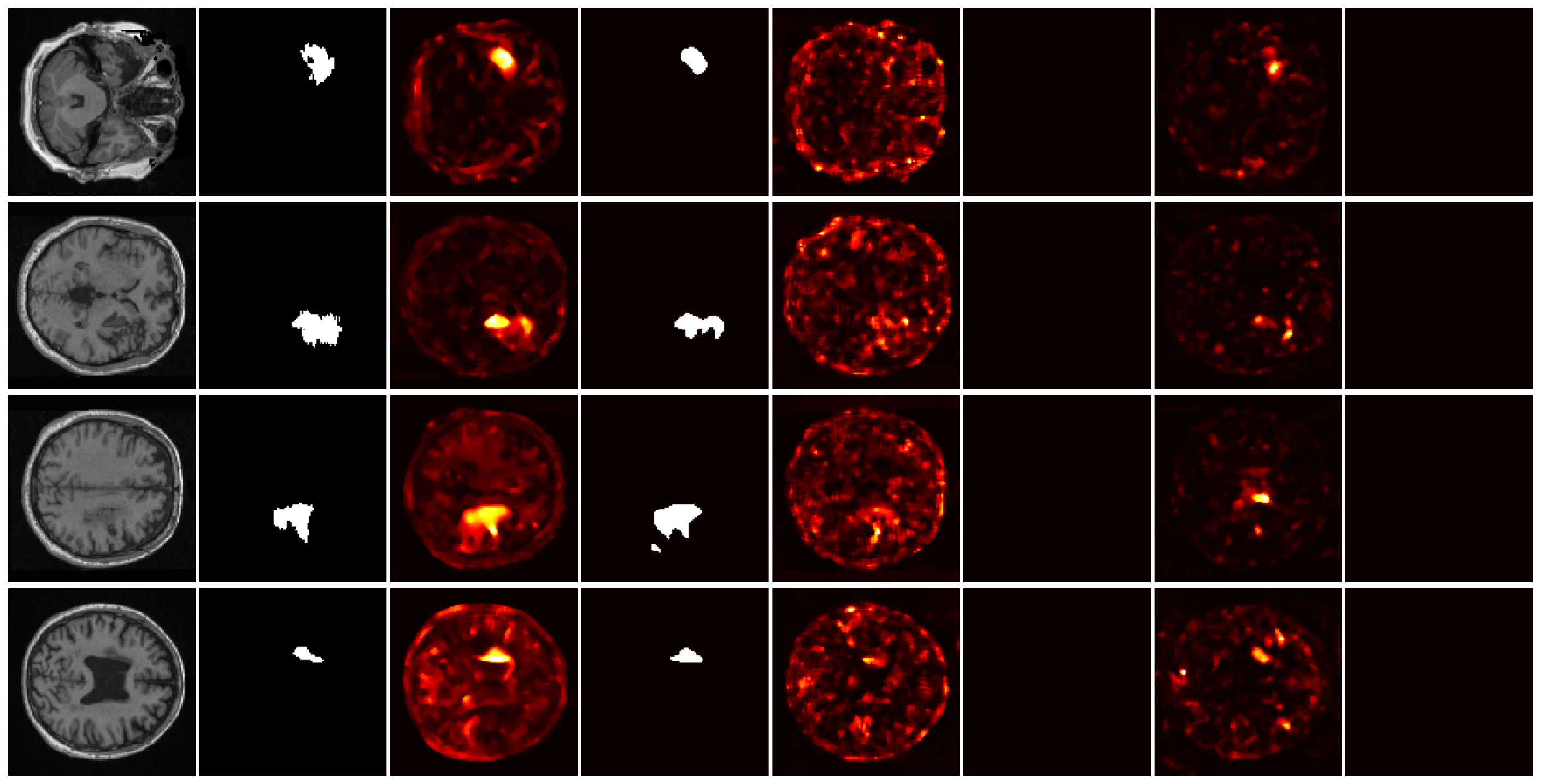}};
       % Adding labels on the top side
       % Adding labels on the top side
       \node at (-8.3,4.40) {Input};
       \node at (-6.30,4.40) {GT};
       \node at (-3.4,4.40) {Ours (DDIM)};
       \node at (0.6,4.40) {DDIM clf-free};
       \node at (4.3,4.40) {Ano DDPM (S)};
     \end{tikzpicture}
     \label{fig:qualitative_atlas_app}
   \end{subfigure}
   \caption{Qualitative Comparison of Anomaly Maps and Segmentation. (a) From the BraTS21 dataset and (b) from the ATLAS v2.0 dataset. The first column displays the original input images, and the second column shows the corresponding ground truth for anomaly segmentation. Subsequent columns present the anomaly maps and segmentation results obtained using our method, AnoFPDM with the DDIM setting, alongside those from the second and third best comparative methods. Each row represents a different sample.}
   \label{fig:qualitative_app}
 \end{figure*}

We provide more qualitative results for the BraTS21 and ATLAS v2.0 datasets in \cref{fig:qualitative_app}. It further shows the effectiveness of our method in detecting anomalies and segmenting them. The signal strength of the anomalies is enhanced by the aggregation of SAMs, leading to more accurate segmentation results.

\section{Postprocessing for Segmentation}
After we obtain the anomaly map, we apply a median filter \cite{kascenas2022denoising} with kernel size 5 to effectively enhance the performance. Then, we apply the connected component filter to remove the small connected components which is regarded as noise. We apply the same postprocessing to all methods for fair comparison.

\section{Discussion and Limitations}
All methods showed better results on the BraTS21 dataset than on the ATLAS v2.0 dataset. This disparity arises because DMs are more adept at identifying anomalies that exhibit significant frequency differences, such as tumors on FLAIR MRI, compared to the surrounding healthy tissue. In this case, the difference of healthy and unhealthy distribution is easier to be captured by DMs. In contrast, the ATLAS v2.0 dataset, which consists of T1 MRI, presents more challenging scenarios for anomaly detection due to the subtle frequency differences between healthy and unhealthy regions. During the inference stage, the implicit classifier struggles to accurately capture the anomalous regions, contributing to the less consistent signal of anomalous regions in the SAMs. This inconsistency can lead to the mixing of signals from falsely detected healthy regions, resulting in lower detection accuracy.

Our selection method is specifically designed for the weakly-supervised setting, where unhealthy samples are available for training the guided diffusion model. In unsupervised settings, the unguided diffusion model is typically trained only on healthy samples and evaluated on unhealthy samples. In this scenario, the unguided forward process (UFP) of samples through the diffusion model is not possible, which is a crucial aspect of our method. We leave the exploration of unsupervised settings for future work.

\end{document}